\documentclass{article}
\usepackage{spconf,amsmath,graphicx}
\usepackage{array,multirow}
\usepackage{arydshln}
\usepackage{adjustbox}
\usepackage{amsthm}
\usepackage{booktabs}
\usepackage{bbding}

\title{Over-Sampling Strategy in Feature Space for Graphs based\\
Class-imbalanced Bot Detection}
\name{Shuhao Shi$^{\dag}$, Kai Qiao$^{\dag}$, Jie Yang, Baojie Song, Jian Chen, Bin Yan$^{*}$\thanks{*Corresponding Author. This work was supported by the National Key Research and Development Project of China under Grants 2020YFC1522002.}}
\address{PLA Strategiy Support Forces Information Engineering University, Zhengzhou, China}
\begin{document}
%
\maketitle
\begin{abstract}
The presence of a large number of bots in Online Social Networks (OSN) leads to undesirable social effects. Graph neural networks (GNNs) are effective in detecting bots as they utilize user interactions. However, class-imbalanced issues can affect bot detection performance. To address this, we propose an over-sampling strategy for GNNs (OS-GNN) that generates samples for the minority class without edge synthesis. First, node features are mapped to a feature space through neighborhood aggregation. Then, we generate samples for the minority class in the feature space. Finally, the augmented features are used to train the classifiers. This framework is general and can be easily extended into different GNN architectures. The proposed framework is evaluated using three real-world bot detection benchmark datasets, and it consistently exhibits superiority over the baselines.
\end{abstract}

\section{Introduction}
Online Social Networks (OSN) have been plagued by bots and malicious accounts, which has caused negative social effects~\cite{article01,article02}. To address this issue, researchers have proposed various bot detection methods. Currently, the most effective bot detection methods~\cite{article17,article21,article22,article32} use Graph Neural Networks (GNNs) to exploit user relationships.

\par In bot detection, users are classified as either bots or humans. The imbalance ratio $\rho=\frac{M}{N}$ is used to measure class imbalance, where $N$ and $M$ represent majority and minority class sample numbers, respectively. Table \ref{tb:statistics} summarizes the existing Twitter bot detection datasets, which mostly have imbalanced bot and human user distributions.

\begin{table}
\caption{Distribution of bot detection datasets class. Bot detection generally suffers from class-imbalance issue. Among all bot detection datasets, only TwiBot-22, MGTAB, TwiBot-20, and Crecsi-15 contain graph structures.}
\begin{center}
\begin{adjustbox}{width=0.725\linewidth}
  \begin{tabular}{l|c|c|c|c}
  \toprule
  \multicolumn{1}{c|}{Dataset} &\multicolumn{1}{c|}{Bot} &\multicolumn{1}{c|}{Human} &\multicolumn{1}{c|}{$\rho$} &\multicolumn{1}{c}{Graph}\\
  \hline
  \rule{0pt}{2.5ex}
  TwiBot-22~\cite{article32}  & 139,943 & 860,057 & 0.163  &\Checkmark\\
  \rule{0pt}{0.5ex}
  Midterm-18~\cite{article34} & 42,446  & 8,092   & 0.191  &\XSolid\\
  \rule{0pt}{0.5ex}
  Cresci-17~\cite{article01}  & 10,894  & 3,474   & 0.319  &\XSolid\\
  \rule{0pt}{0.5ex}
  MGTAB~\cite{article25}      & 2,830   & 7,554   & 0.375  &\Checkmark\\
  \rule{0pt}{0.5ex}
  Kaiser~\cite{article33}     & 875     & 499     & 0.570  &\XSolid\\
  \rule{0pt}{0.5ex}
  Cresci-15~\cite{article24}  & 3,351   & 1,950   & 0.582  &\Checkmark\\
  \rule{0pt}{0.5ex}
  Gilani-17~\cite{article35}  & 1,090   & 1,394   & 0.782  &\XSolid\\
  \rule{0pt}{0.5ex}
  TwiBot-20~\cite{article23}  & 6,589   & 5,237   & 0.795  &\Checkmark\\
  \bottomrule
  \end{tabular}
  \label{tb:statistics}
  \end{adjustbox}
  \vspace{-0.05in}
  \end{center}
\end{table}

\par
\begin{figure}[h]
\vspace{-0.10in}
   \centering
   \begin{minipage}[b]{0.375\textwidth}
    \includegraphics[width=\textwidth]{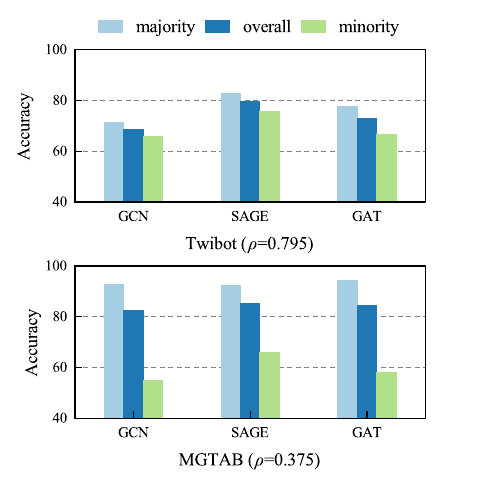}
  \end{minipage}
  \vspace{-0.1cm}
  \caption{\small Test accuracy on MGTAB and TwiBot-20 dataset. The poor performance of minority class not reflected in overall performance.}
  \vspace{-0.1cm}
  \label{fig:fp}
\end{figure}

 Graph-based bot detection methods have not adequately addressed the issue of class imbalance. When the training data is imbalanced, the model may struggle to learn enough features from the minority class, resulting in a tendency to classify most samples as belonging to the majority class. Figure~\ref{fig:fp} shows the results of bot detection using GNNs on the MGTAB~\cite{article25} and TwiBot-20~\cite{article23} datasets. Despite high overall accuracy, there is a significant bias towards predicting the majority class, with the minority class being much less accurately classified~\cite{article29}. This is especially the case when the degree of imbalance is large.

In this paper, we propose Over-Sampling Strategy in Feature Space for Graphs Neural Networks (OS-GNN) to address class imbalance in GNN models in bot detection. Our approach generates synthetic features for the minority class by oversampling in the feature space, without synthesizing node edges. We then use the rebalanced feature matrix for classification. Our method outperforms previous methods such as GraphSMOTE~\cite{article06} and GraphEns~\cite{article07} on various bot detection benchmark datasets with different GNN architectures.

\section{Preliminary}
\label{sec:Preliminary}
\subsection{Notations and Problem Definition}
The distribution of class scale for bot and human are imbalanced, as Table \ref{tb:statistics} shown, resulting in GNN classifiers' bias towards the majority class.

\par Let $\mathrm{G}=(\mathrm{V}, \mathrm{E}, \mathrm{H})$ denote a social graph, where $\mathrm{V}=\left\{v_{1}, v_{2}, \ldots, v_{N}\right\}$ is a set of users and $\mathrm{E}$ is a set of relationships between them. The relationships include friends and followers. $\mathrm{H} \in \mathrm{R}^{N \times d}$ is the feature matrix, where $d$ is the dimension of the node attribute. $\mathrm{Y} \in \mathrm{R}^{n}$ is the label for nodes in $\mathrm{G}$. We denote the set of labeled nodes as $\mathrm{V}{L}$ and the set of synthetic nodes as $\mathrm{V}{S}$. In bot detection, labels are obtained through manual annotation, which can be costly. Thus, $\left|\mathrm{V}_{L}\right|<<|\mathrm{V}|$ is typical. Our goal is to generate nodes for the minority class to improve accuracy on the minority class and overall model accuracy in class-imbalanced graphs.

\begin{equation}
\label{eq:Definition}
f\left(\mathrm{~V}_{L}, \mathrm{~V}_{S}, \mathrm{H}\right) \rightarrow \mathrm{Y}
\end{equation}

\subsection{Graph Neural Networks} GNNs learn node representations according to the graph structure, and the process can be formulated as:

\begin{equation}
\label{eq:GNN}
\resizebox{0.95\linewidth}{!}{$
h_{v}^{(l)}=\text{COMBINE}\left(h_{v}^{(l)}, \text{AGGREGATE}\left(\left\{h_{u}^{(l-1)}, \forall u \in N_{v}\right\}\right)\right)
$},
\end{equation}

where $h_{v}^{l}$ denotes feature of node $v$ at $l$-th GNN layer and $N_{v}$ represent the neighbor set of node $v$ ; $h_{v}^{(0)}$ is initialized with node attribute $h_{v}$. $\text{AGGREGATE}(\cdot)$ and $\text{COMBINE}(\cdot)$ represent neighbor aggregation and combination functions, respectively. The aggregation function usually needs to be differentiable and permutation invariant.

\section{Proposed Method}
\label{sec:Proposed_Method}
\subsection{Motivation}
Current over-sampling methods for GNNs~\cite{article06,article15} create nodes in either the original feature space or the embedding space. However, the newly generated nodes lack connectivity relationships (edges), which must be constructed after over-sampling. This results in an approximation of the true distribution of edges and increases noise in the synthetic data, ultimately impacting the GNNs' performance.

OS-GNN addresses this issue by generating embeddings for minority class nodes in feature space to balance the distribution of different classes.

\begin{figure*}[h]
  \centering
   \begin{minipage}{0.75\textwidth}
    \centerline{\includegraphics[width=\textwidth]{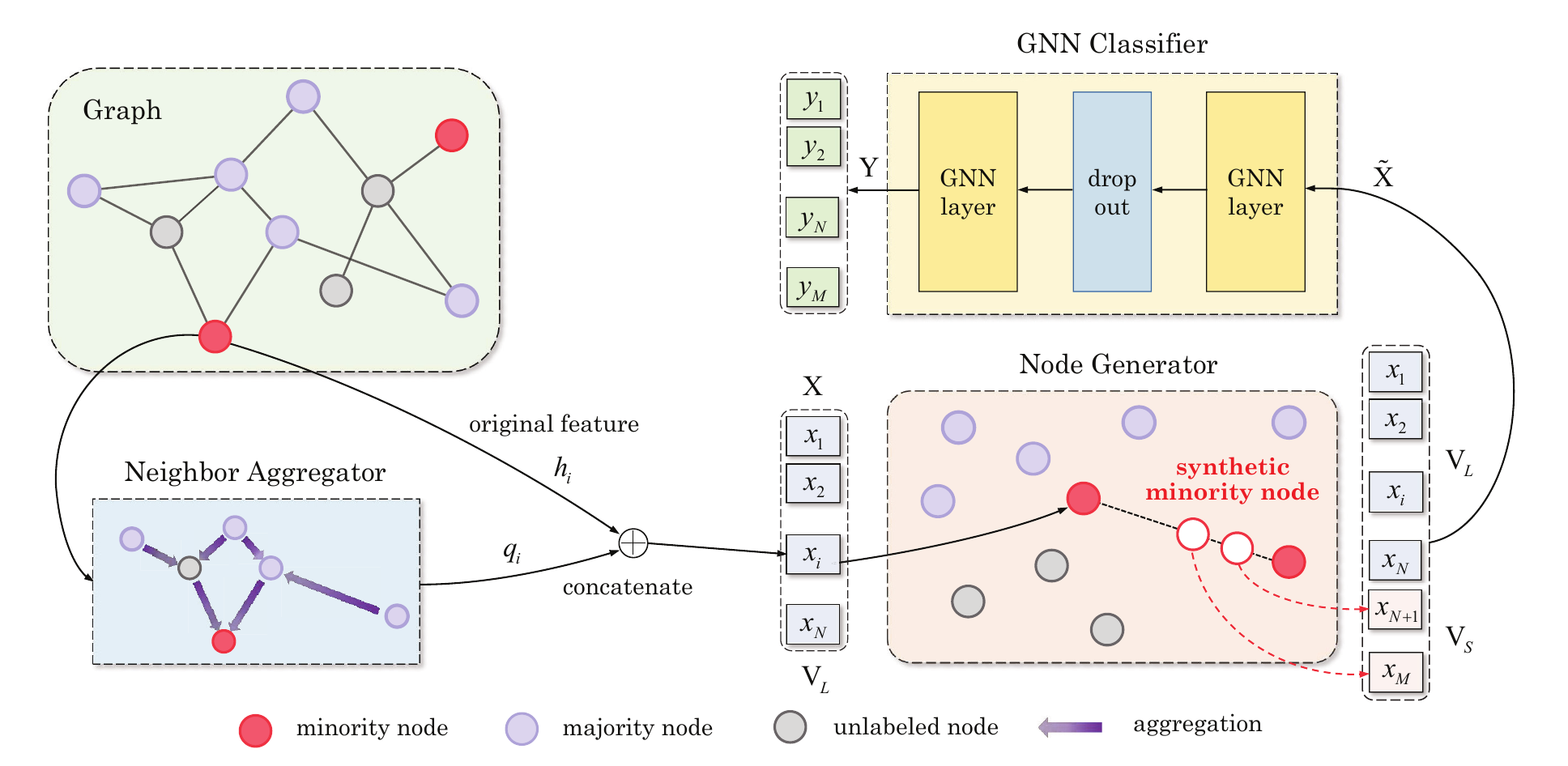}}
   \end{minipage}
   \vspace{-0.1cm}
  \caption{Overview of the OS-GNN. $v_{i}$ is a node sampled from the minority class. $x_{i}$ is acquired by concatenating original feature $h_{i}$ and embedding $q_{i}$ obtained by neighbor aggregator. The augmented $\tilde{\mathrm{X}}$, obtained by over-sampling $\mathrm{X}$, is fed into the GNN classifier.}
  \label{fig:schematic}
  \vspace{-0.2cm}
\end{figure*}

\subsection{Neighbor Aggregator}
\label{sec:Neighbor_Aggregator}
Neighborhood aggregation obtains node embeddings using neighborhood information. The original attribute of node $v_{i}$ is $h_{i}$ and the representation with neighborhood information $q_{i}$ is obtained through the neighbor aggregator. We use 2-hop neighborhood aggregation to prevent over-smoothing and over-fitting, i.e., $q_{i}=h_{i}^{2}$, and formulate the neighbor aggregation as in Eq. (\ref{eq:GNN}).

\par As the original graph is imbalanced, some bot nodes may be largely connected to human nodes. Neighborhood aggregation may make the embeddings of these bot nodes similar to human nodes. To preserve the original node information, we concatenate $q_{i}$ with $h_{i}$ to obtain $x_{i}=\left[q_{i} \| h_{i}\right]$, the representation of node $v_{i}$.

\subsection{Synthetic Node Generator}
After obtaining $\mathrm{X}=\left[x_{i}, 0 \leq i \leq N\right]$, we apply SMOTE~\cite{article26} to generate synthetic minority nodes. In this way, we can make the distribution of different classes more balanced, making the trained classifier perform better on minority class samples initially under-represented.

We use the hyper-parameter $\omega$, over-sampling scale, to control the amount of nodes to be generated for minority class. For minority class node $v_{i}$, let $\Gamma(\cdot)$ denote the set of k-nearest neighbors measured by Euclidean distance in the feature space. A random node $v_{u}$, with the same labels as $v_{i}$, is selected from $\Gamma\left(v_{i}\right)$. A random point on the line between $x_{u}$ and $x_{i}$ is chosen as $x_{k}$:

\begin{equation}
\label{eq:SMOTE}
x_{k}=(1-\delta) \cdot x_{i}+\delta \cdot x_{u},
\end{equation}

where $x_{k}$ is the embedding of the virtual node $v_{k}$ in the feature space. $\delta$ is a random variable, following uniform distribution, ranges from 0 to 1. $v_{k}$ has the same labels as $v_{i}$ and $v_{u}$. Thus, we obtain labeled synthetic minority class samples that equalize the sample distribution.

\subsection{GNN Classifier}
Let $\tilde{\mathrm{X}}$ be the augmented node representations in the feature space, concatenating $\mathrm{X}$ with the embedding of the synthetics nodes. As the embeddings already contain neighbourhood information, the generation of edges for the synthetic nodes is not required. For the GCN classifier, the formula is shown as:

\begin{equation}
\label{eq:GCN}
\hat{\mathrm{Y}}=\operatorname{softmax}\left(\mathrm{A} \sigma\left(\mathrm{A}\mathrm{H} \mathrm{W}^{(1)}\right) \mathrm{W}^{(2)}\right).
\end{equation}

Use the identity matrix $\mathrm{E}$ as the input adjacency matrix $\mathrm{A}$ of GNN classifier $f(\cdot)$. The labels and the augmented embeddings matrix $\tilde{\mathrm{X}}$ are input into $f$ for training. The formula for our OS-GCN, using GCN as backbone, is shown as Eq. (\ref{eq:OS_GCN}).

\begin{equation}
\label{eq:OS_GCN}
\hat{\mathrm{Y}}=\operatorname{softmax} \left(\sigma\left(\tilde{\mathrm{X}} \mathrm{W}^{(1)}\right). \mathrm{W}^{(2)}\right)
\end{equation}

The final objective function of GNN classifier can be written as:

\begin{equation}
\label{eq:objective_function}
\mathcal{L}\left(y^{\prime}\right)=\sum_{v_{i} \in V^{L}} \operatorname{loss}\left(y_{i}, y_{i}^{\prime}\right)+\lambda \sum_{v_{j} \in V^{s}} \operatorname{loss}\left(y_{j}, y_{j}^{\prime}\right),
\end{equation}

where $y_{i}^{\prime}$ denotes predicted label of node $v_{i}$, and $\lambda$ is the parameter ranging from 0 to 1. The loss function measures the supervised loss between real and predicted labels. Different GNN models can be adopted for classifier, and the entire framework is efficient and easy to implement.

\begin{table*}[t]
\vspace{-0.1cm}
\caption{Experimental results of our OS-GNN and other baselines on three bot detection benchmark datasets. We report averaged accuracy (Acc), F1-macro score (F1), and balanced accuracy (bAcc) and with the standard errors for 5 repetitions on three representative GNN architectures.}
\vspace{-0.1cm}
\begin{center}
\begin{scriptsize}
\setlength{\columnsep}{1pt}%
\begin{adjustbox}{width=0.905\linewidth}
\begin{tabular}{@{\extracolsep{1pt}}rlccc|ccc|ccc@{}}
\toprule
 & \multirow{2}{*}{\textbf{Method}} & \multicolumn{3}{c}{TwiBot-20 ($\rho=0.795$, $\omega=0.258$)} & \multicolumn{3}{c}{Cresci-15 ($\rho=0.582$, $\omega=0.718$)} & \multicolumn{3}{c}{MGTAB ($\rho=0.375$, $\omega=1.667$)}  \\
\cline{3-11}
\rule{0pt}{2.2ex}
& & Acc & F1 &bAcc & Acc & F1 &bAcc & Acc & F1 &bAcc\\
\cline{2-11}
\rule{0pt}{2.5ex}
\multirow{8}{*}{\rotatebox{90}{GCN}}
& Vanilla    & \scriptsize{68.76} \tiny{$\pm 0.60$}& 68.30 \tiny{$\pm 0.51$}& 68.29 \tiny{$\pm 0.62$}
             & 96.50 \tiny{$\pm 0.36$}& 96.20 \tiny{$\pm 0.42$}& 95.95 \tiny{$\pm 0.53$}
             & 82.69 \tiny{$\pm 0.76$}& 74.85 \tiny{$\pm 1.32$}& 72.32 \tiny{$\pm 1.29$}
             \\
\cdashline{2-11}
\rule{0pt}{2.5ex}
& Focal loss & 69.83 \tiny{$\pm 0.95$}& 69.68 \tiny{$\pm 0.86$}& 68.37 \tiny{$\pm 0.79$}
             & 96.61 \tiny{$\pm 0.31$}& 96.27 \tiny{$\pm 0.82$}& 96.23 \tiny{$\pm 0.65$}
             & 84.43 \tiny{$\pm 0.68$}& 78.39 \tiny{$\pm 1.17$}& 76.31 \tiny{$\pm 1.54$}
             \\
& CB loss    & 69.94 \tiny{$\pm 0.55$}& 69.81 \tiny{$\pm 0.53$}& 68.36 \tiny{$\pm 0.57$}
             & 94.03 \tiny{$\pm 2.08$}& 93.77 \tiny{$\pm 4.03$}& 94.25 \tiny{$\pm 3.86$}
             & 84.12 \tiny{$\pm 0.65$}& 80.14 \tiny{$\pm 0.91$}& 80.60 \tiny{$\pm 1.15$}
             \\
& DR-GCN     & 76.40 \tiny{$\pm 1.05$}& 75.50 \tiny{$\pm 1.34$}& 76.23 \tiny{$\pm 0.72$}
             & 93.96 \tiny{$\pm 2.84$}& 93.64 \tiny{$\pm 3.37$}& 94.01 \tiny{$\pm 2.20$}
             & 83.23 \tiny{$\pm 2.93$}& 72.58 \tiny{$\pm 5.10$}& 76.32 \tiny{$\pm 3.25$}
             \\
& RA-GCN     & 73.65 \tiny{$\pm 0.65$}& 73.14 \tiny{$\pm 0.49$}& 72.08 \tiny{$\pm 1.24$}
             & 96.67 \tiny{$\pm 0.11$}& 96.40 \tiny{$\pm 0.12$}& 95.74 \tiny{$\pm 0.31$}
             & 82.13 \tiny{$\pm 0.85$}& 78.02 \tiny{$\pm 0.85$}& 78.60 \tiny{$\pm 1.37$}
             \\
& GraphSmote & 76.40 \tiny{$\pm 0.79$}& 71.50 \tiny{$\pm 1.47$}& 75.25 \tiny{$\pm 1.01$}
             & 96.56 \tiny{$\pm 0.22$}& 96.26 \tiny{$\pm 0.27$}& 95.98 \tiny{$\pm 0.31$}
             & 83.28 \tiny{$\pm 0.61$}& 79.21 \tiny{$\pm 0.73$}& 81.96 \tiny{$\pm 1.32$}
             \\
& GraphEns   & 76.12 \tiny{$\pm 0.79$}& 74.39 \tiny{$\pm 0.38$}& 76.13 \tiny{$\pm 0.76$}
             & 96.61 \tiny{$\pm 0.17$}& 96.31 \tiny{$\pm 0.19$}& 96.17 \tiny{$\pm 0.23$}
             & 82.83 \tiny{$\pm 0.85$}& 77.21 \tiny{$\pm 2.43$}& 78.63 \tiny{$\pm 1.43$}
             \\
\cline{2-11}
\rule{0pt}{2.5ex}
& Ours       & \textbf{83.44} \tiny{$\pm 0.40$}& \textbf{83.18} \tiny{$\pm 0.35$}& \textbf{83.12} \tiny{$\pm 0.24$}
             & \textbf{96.73} \tiny{$\pm 0.30$}& \textbf{96.46} \tiny{$\pm 0.18$}& \textbf{96.43} \tiny{$\pm 0.19$}
             & \textbf{85.84} \tiny{$\pm 0.92$}& \textbf{83.27} \tiny{$\pm 0.80$}& \textbf{85.81} \tiny{$\pm 0.33$}\\
\cline{2-11}
\noalign{\vskip\doublerulesep
         \vskip-\arrayrulewidth} \cline{2-11}
\rule{0pt}{2.5ex}
\multirow{8}{*}{\rotatebox{90}{SAGE}}
& Vanilla    & 79.65 \tiny{$\pm 0.59$}& 79.31 \tiny{$\pm 0.56$}& 78.85 \tiny{$\pm 0.79$}
             & 96.41 \tiny{$\pm 0.29$}& 95.41 \tiny{$\pm 0.33$}& 96.04 \tiny{$\pm 0.24$}
             & 85.34 \tiny{$\pm 0.49$}& 81.70 \tiny{$\pm 0.83$}& 79.33 \tiny{$\pm 1.67$}
             \\
\cdashline{2-11}
\rule{0pt}{2.5ex}
& Focal loss & 78.02 \tiny{$\pm 1.57$}& 77.69 \tiny{$\pm 1.53$}& 78.24 \tiny{$\pm 1.49$}
             & 96.22 \tiny{$\pm 0.31$}& \textbf{95.89} \tiny{$\pm 0.31$}& 95.61 \tiny{$\pm 0.55$}
             & 85.84 \tiny{$\pm 0.51$}& 82.20 \tiny{$\pm 0.63$}& 82.17 \tiny{$\pm 1.10$}
             \\
& CB loss    & 78.49 \tiny{$\pm 0.48$}& 77.23 \tiny{$\pm 0.45$}& 77.12 \tiny{$\pm 0.70$}
             & 96.04 \tiny{$\pm 0.32$}& 95.73 \tiny{$\pm 0.35$}& 95.70 \tiny{$\pm 0.57$}
             & 86.02 \tiny{$\pm 0.76$}& 82.76 \tiny{$\pm 0.87$}& 83.92 \tiny{$\pm 0.75$}
             \\
& DR-GCN     & 79.87 \tiny{$\pm 0.80$}& 78.50 \tiny{$\pm 0.66$}& 79.01 \tiny{$\pm 0.55$}
             & 95.52 \tiny{$\pm 1.02$}& 95.64 \tiny{$\pm 2.26$}& 94.87 \tiny{$\pm 1.18$}
             & 85.61 \tiny{$\pm 0.94$}& 81.15 \tiny{$\pm 1.20$}& 78.82 \tiny{$\pm 1.16$}
             \\
& RA-GCN     & 77.65 \tiny{$\pm 0.44$}& 78.14 \tiny{$\pm 0.67$}& 76.65 \tiny{$\pm 1.03$}
             & 96.32 \tiny{$\pm 0.27$}& 96.11 \tiny{$\pm 0.42$}& \textbf{96.16} \tiny{$\pm 0.24$}
             & 85.23 \tiny{$\pm 0.85$}& 78.02 \tiny{$\pm 0.85$}& 78.06 \tiny{$\pm 1.12$}
             \\
& GraphSmote & 80.25 \tiny{$\pm 0.74$}& 79.14 \tiny{$\pm 0.95$}& 79.35 \tiny{$\pm 0.54$}
             & 96.34 \tiny{$\pm 0.72$}& 95.82 \tiny{$\pm 0.96$}& 94.28 \tiny{$\pm 2.17$}
             & 85.51 \tiny{$\pm 0.62$}& 81.15 \tiny{$\pm 0.70$}& 78.22 \tiny{$\pm 0.75$}
             \\
& GraphEns   & 79.85 \tiny{$\pm 0.37$}& 79.42 \tiny{$\pm 0.46$}& 78.83 \tiny{$\pm 0.61$}
             & 96.04 \tiny{$\pm 0.28$}& 95.73 \tiny{$\pm 0.37$}& 94.56 \tiny{$\pm 1.34$}
             & 86.19 \tiny{$\pm 0.54$}& 81.45 \tiny{$\pm 0.58$}& 81.04 \tiny{$\pm 0.63$}
             \\
\cline{2-11}
\rule{0pt}{2.5ex}
& Ours       & \textbf{81.32} \tiny{$\pm 0.65$}& \textbf{81.03} \tiny{$\pm 0.61$}& \textbf{81.00} \tiny{$\pm 0.65$}
             & \textbf{96.42} \tiny{$\pm 0.33$}& 95.62 \tiny{$\pm 0.35$}& 96.06 \tiny{$\pm 0.61$}
             & \textbf{86.21} \tiny{$\pm 0.35$}& \textbf{82.81} \tiny{$\pm 0.24$}& \textbf{84.36} \tiny{$\pm 0.44$}\\

\cline{2-11}
\noalign{\vskip\doublerulesep
         \vskip-\arrayrulewidth} \cline{2-11}
\rule{0pt}{2.5ex}
\multirow{8}{*}{\rotatebox{90}{GAT}}
& Vanilla    & 72.80 \tiny{$\pm 0.11$}& 72.31 \tiny{$\pm 0.27$}& 71.57 \tiny{$\pm 0.88$}
             & 96.49 \tiny{$\pm 0.15$}& 96.18 \tiny{$\pm 0.30$}& 95.86 \tiny{$\pm 0.39$}
             & 84.46 \tiny{$\pm 1.13$}& 80.47 \tiny{$\pm 1.29$}& 79.35 \tiny{$\pm 1.58$}
             \\
\cdashline{2-11}
\rule{0pt}{2.5ex}
& Focal loss & 71.12 \tiny{$\pm 1.46$}& 71.66 \tiny{$\pm 1.40$}& 71.22 \tiny{$\pm 1.03$}
             & 96.43 \tiny{$\pm 0.26$}& 96.11 \tiny{$\pm 0.29$}& 95.65 \tiny{$\pm 0.47$}
             & 83.77 \tiny{$\pm 1.15$}& 78.60 \tiny{$\pm 0.99$}& 77.68 \tiny{$\pm 1.57$}
             \\
& CB loss    & 72.13 \tiny{$\pm 0.73$}& 71.52 \tiny{$\pm 0.66$}& 71.33 \tiny{$\pm 1.80$}
             & 96.47 \tiny{$\pm 0.32$}& 96.18 \tiny{$\pm 0.34$}& 95.98 \tiny{$\pm 0.42$}
             & 84.58 \tiny{$\pm 0.72$}& 81.12 \tiny{$\pm 0.70$}& 82.35 \tiny{$\pm 1.04$}
             \\
& DR-GCN     & 76.40 \tiny{$\pm 1.05$}& 75.50 \tiny{$\pm 1.34$}& 75.87 \tiny{$\pm 1.45$}
             & 93.16 \tiny{$\pm 4.84$}& 92.64 \tiny{$\pm 5.03$}& 92.01 \tiny{$\pm 2.36$}
             & 83.23 \tiny{$\pm 2.93$}& 74.58 \tiny{$\pm 3.10$}& 76.68 \tiny{$\pm 1.57$}
             \\
& RA-GCN     & 76.17 \tiny{$\pm 0.65$}& 75.47 \tiny{$\pm 1.23$}& 75.81 \tiny{$\pm 0.85$}
             & 94.15 \tiny{$\pm 0.25$}& 93.80 \tiny{$\pm 0.25$}& 94.32 \tiny{$\pm 0.23$}
             & 84.76 \tiny{$\pm 0.65$}& 80.28 \tiny{$\pm 2.51$}& 78.98 \tiny{$\pm 1.02$}
             \\
& GraphSmote & 77.42 \tiny{$\pm 0.61$}& 76.34 \tiny{$\pm 1.05$}& 76.39 \tiny{$\pm 1.63$}
             & 96.55 \tiny{$\pm 0.27$}& 96.38 \tiny{$\pm 0.30$}& 95.56 \tiny{$\pm 0.49$}
             & 84.92 \tiny{$\pm 1.35$}& 81.13 \tiny{$\pm 2.23$}& 76.84 \tiny{$\pm 1.72$}
             \\
& GraphEns   & 76.17 \tiny{$\pm 0.25$}& 75.47 \tiny{$\pm 0.32$}& 75.41 \tiny{$\pm 0.53$}
             & 96.57 \tiny{$\pm 0.26$}& \textbf{96.45} \tiny{$\pm 0.33$}& 94.95 \tiny{$\pm 0.45$}
             & 85.06 \tiny{$\pm 0.65$}& 80.82 \tiny{$\pm 1.21$}& 74.32 \tiny{$\pm 1.14$}
             \\
\cline{2-11}
\rule{0pt}{2.5ex}
& Ours       & \textbf{82.49} \tiny{$\pm 0.42$}& \textbf{82.30} \tiny{$\pm 0.37$}& \textbf{82.41} \tiny{$\pm 0.25$}
             & \textbf{96.65} \tiny{$\pm 0.36$}& 96.38 \tiny{$\pm 0.39$}& \textbf{96.35} \tiny{$\pm 0.40$}
             & \textbf{86.75} \tiny{$\pm 0.74$}& \textbf{85.39} \tiny{$\pm 0.71$}& \textbf{87.18} \tiny{$\pm 0.50$}\\
\bottomrule

\end{tabular}
\end{adjustbox}
\end{scriptsize}
\end{center}
\label{tb:main_results}
\end{table*}

\section{Experiment}
\label{sec:Experiment}
\subsection{Experimental Settings}
\noindent \textbf{Datasets}
We evaluate models on three Twitter bot detection datasets that with graph structures: Cresci-15~\cite{article24}, TwiBot-20~\cite{article23}, MGTAB~\cite{article25}. We conduct a 1:1:8 random partition as training, validation, and test set for all datasets, and use the most commonly used follower and friend relationships during evaluation.

\noindent \textbf{Evaluation Metrics}
We adopt three criteria: Accuracy (Acc), F1-macro scores (F1), as well as balanced accuracy (bAcc) for the class-imbalanced node classification. To further evaluate performance on minority class data, we adopt two criteria: true positive rate (TPR) and minority class accuracy (Minor Acc).

\noindent \textbf{Configurations}
All experiments were conducted on a server with 9 GPUs (RTX TITAN). All models were trained using the Adam optimization algorithm and 2 layers for GNN. The learning rate was initialized to 1$e$-3 with weight decay being 5$e$-4. Models were trained until convergence, with a maximum epoch of 500. The minority class samples were generated to match the majority class using $\omega$ set to $\frac{N-M}{M}$ ($\frac{1-\rho}{\rho}$). Hyper-parameter $\lambda$ was set to 0.8 for optimal accuracy and TPR. $\beta \in\{0.9,0.99,0.999,0.9999\}$ for CB loss and $\alpha$ was set to 0.25 for Focal loss. Focal loss and CB loss both used $\gamma$ set to 2.

\subsection{Baseline methods}
To validate our method, we compare our model with a variety of baseline methods using different rebalance methods. We consider seven baseline methods, including vanilla (cross entropy), Focal loss~\cite{article14}, Class Balanced (CB) loss~\cite{article13}, DR-GCN~\cite{article08}, RA-GCN~\cite{article05}, GraphSMOTE~\cite{article06}, and GraphENS~\cite{article07}.

\subsection{Main Results}
\label{sec:main_results}
\textbf{Bot detection performance}
We adopt three widely used homogeneous GNNs as backbone: GCN~\cite{article10}, SAGE~\cite{article11}, and GAT ~\cite{article09}. In Table \ref{tb:main_results}, we report average results with standard deviation for the baselines and ours. All OS-GNN with different backbones showed significant improvements compared to the “Vanilla” setting, in which no special algorithm is adopted. In most cases, OS-GNN outperforms all baselines in all datasets, demonstrating our method's effectiveness. Across all datasets, OS-GNN achieved a minor boost on the Cresci-15 dataset. This is due to the unimportance of edges in the Cresci-15 for node classification, reflected in heterogeneous GNNs only slightly outperform homogeneous GNNs. In addition, bot detection on Cresci-15 is a simple task with less negative effects of imbalance.

\begin{figure}[t]
  \centering
   \begin{minipage}[b]{0.21\textwidth}
    \includegraphics[width=1.00\textwidth]{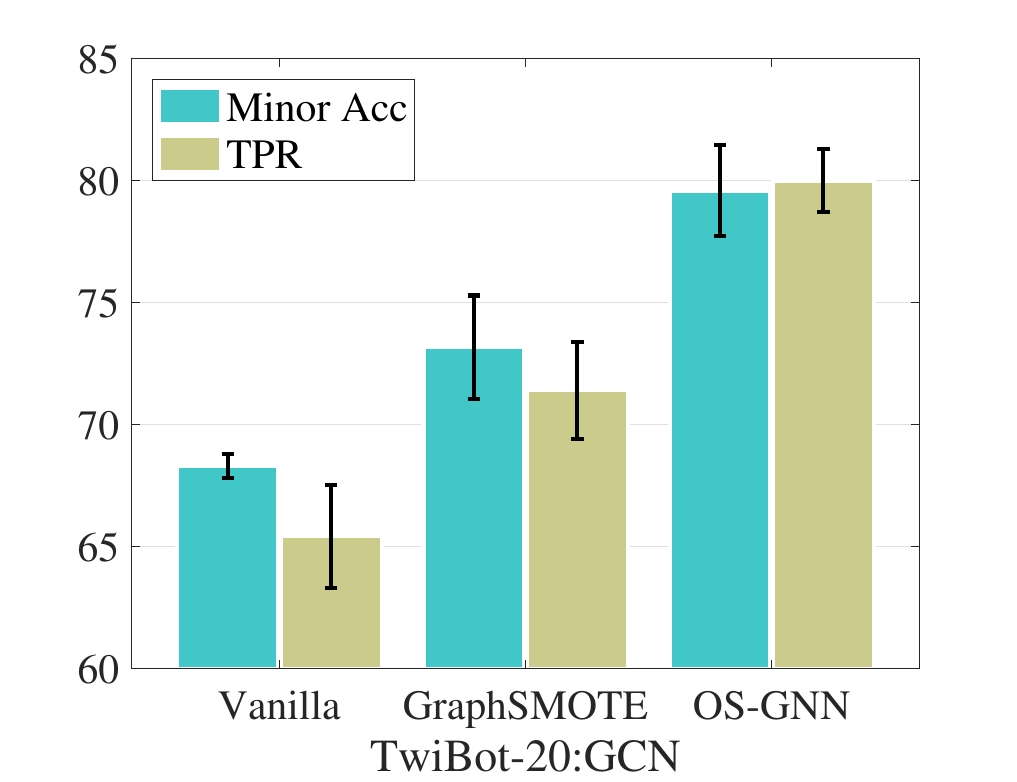}
  \end{minipage}
   \begin{minipage}[b]{0.21\textwidth}
    \includegraphics[width=1.00\textwidth]{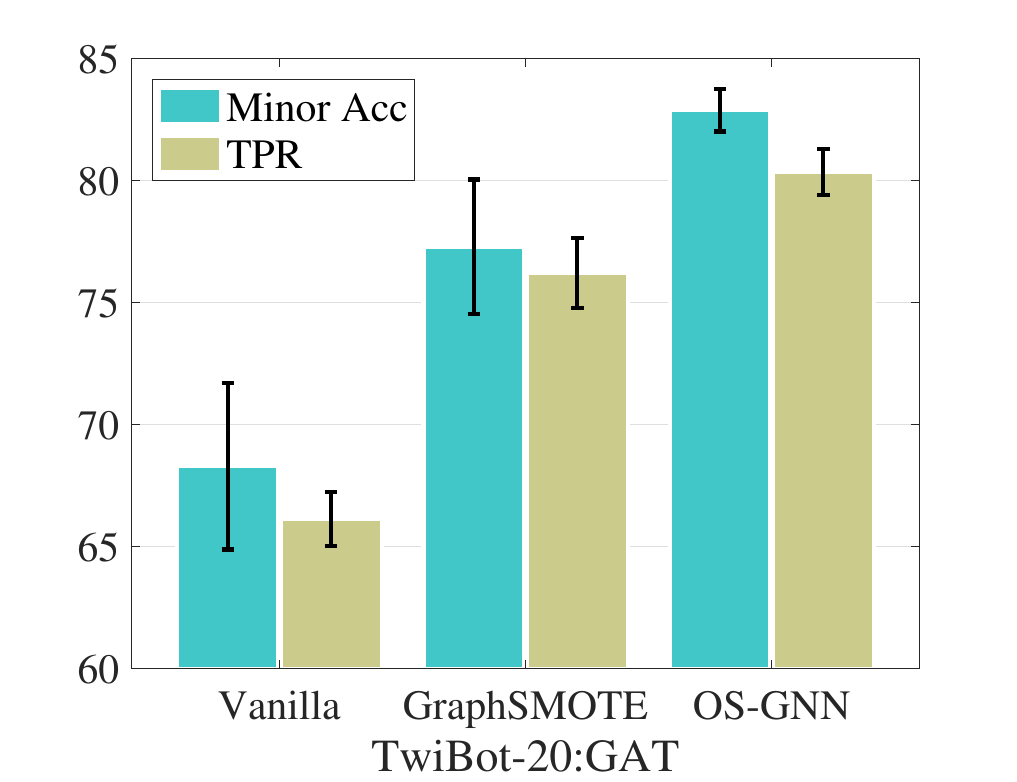}
  \end{minipage}
  \vspace{-0.1cm}
  \caption{Comparison of TPR and Minor Acc on the TwiBot-20 datasets. (a) and (b) use GCN and GAT as backbone model, respectively.}
  \vspace{-0.1cm}
  \label{fig:tpr_minor}
\end{figure}

\begin{figure}[h]
  \centering
   \centering
   \begin{minipage}[b]{0.355\textwidth}
    \includegraphics[width=\textwidth]{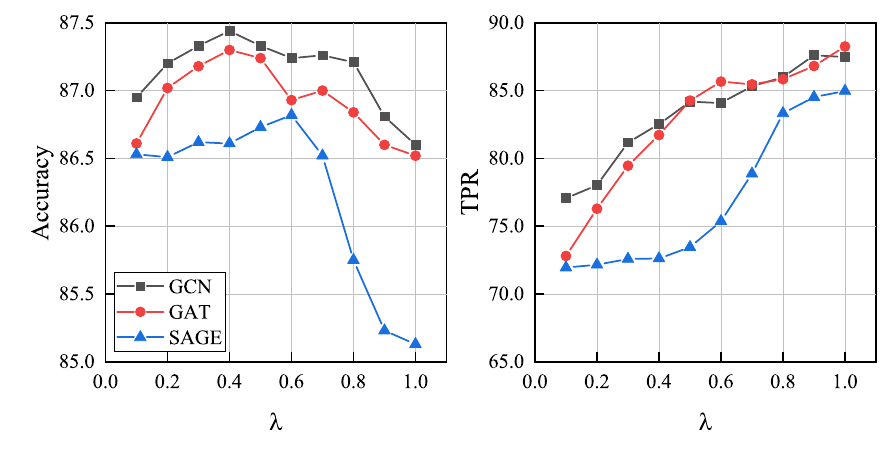}
  \end{minipage}
  \vspace{-0.1cm}
  \caption{\small Sensitivity to hyperparameters $\lambda$. The picture on the left and right shows the accuracy and TPR change of OS-GNN as $\lambda$ increases, respectively.}
  \vspace{-0.1cm}
  \label{fig:sensitivity}
\end{figure}

\noindent \textbf{Improving True Positives Rate}
In Figure \ref{fig:fp}, the minority class classification is low because the model's classification boundary overlaps with the minority class. OS-GNN significantly improves bAcc compared to baseline models, as shown in Table \ref{tb:main_results}. We verify that OS-GNN improves GNN performance by reducing false positives, as seen in the steady increase of TPR and Minor Acc in Figure \ref{fig:tpr_minor}.

\subsection{Parameters Sensitivity Analysis}
We tested OS-GNN with different values of $\lambda$ in Eq. (\ref{eq:objective_function}). We experimented with values from 0.1 to 1.0, running each test five times, and averaging the results shown in Figure \ref{fig:sensitivity}.

Our method is sensitive to $\lambda$. When $\lambda$ increases, accuracy initially improves before decreasing. This is because $\lambda$ controls the weight of synthesized minority class samples in the loss function. As $\lambda$ increases, the model focuses more on the synthesis of minority samples, resulting in a continuous improvement in TPR. However, an excessive $\lambda$ makes the model concentrate too much on synthetic samples, which reduces the performance of majority class samples and ultimately affects the overall classification accuracy.

\section{Conclusions}
In this paper, we introduces Over-Sampling Strategy in Feature Space for Graphs Neural Networks (OS-GNN), an efficient method for tackling the challenge of imbalance in bot detection. Instead of constructing edges, OS-GNN generates embeddings in the feature space for the minority class, reducing noise and improving efficiency. Experimental results on bot detection benchmark datasets demonstrated OS-GNN's effectiveness.

\vfill\pagebreak

\bibliographystyle{IEEEbib}
\bibliography{refs}

\end{document}